\documentclass[11pt]{article}

\usepackage[preprint]{acl}
\usepackage{booktabs}
\usepackage{times}
\usepackage{latexsym}
\usepackage{amsmath}
\usepackage{fvextra}

\DefineVerbatimEnvironment{Prompt}{Verbatim}{
  breaklines=true,
  breakanywhere=true,
  fontsize=\footnotesize,
  frame=single,
  framesep=2mm
}
\usepackage[T1]{fontenc}

\usepackage[utf8]{inputenc}

\usepackage{microtype}

\usepackage{inconsolata}

\usepackage{graphicx}

\title{ConMem: Counterfactual Multi-Scenario Probing for Resolving LLM Memory Misalignment}

\author{
  Shuning Zhang\textsuperscript{1,*},
  Tenghao Ji\textsuperscript{2,*},
  Jingruo Chen\textsuperscript{3}, \\
  \textbf{Eryue Xu\textsuperscript{4},
  Bohao Chu\textsuperscript{5},
  Xin Yi\textsuperscript{1}} \\
  \textsuperscript{1}Tsinghua University \\
  \textsuperscript{2}Independent Researcher \\
  \textsuperscript{3}Cornell University \\
  \textsuperscript{4}University of Illinois Urbana-Champaign \\
  \textsuperscript{5}University of Duisburg-Essen
}

\begin{document}
\maketitle
\begin{abstract}
Long-term memory is critical for autonomous Large Language Model (LLM) agents executing multi-session, long-horizon tasks. However, existing memories often inappropriately store, retain, or retrieve over-generalized, misattributed, or privacy-sensitive information, a problem we term memory misalignment. To study this problem, we construct \textbf{MemAlign}, a benchmark comprising 1,079 curated memory cases grounded in synthetic, validated multi-session conversational contexts covering diverse misalignment taxonomies. We then propose ConMem, an algorithm that reframes retention decisions as prospective counterfactual causal inference. \textbf{ConMem} synthesizes counterfactual interaction scenarios to prospectively audit downstream intervention effects across three aspects: relevant task execution, irrelevant invariance checks, and risk-sensitive boundary shifts. Experiments show that ConMem outperforms other methods, achieving 75.16\% accuracy, 64.15\% balanced accuracy, and a Macro-F1 of 65.31\%, effectively mitigating over-retention biases without sacrificing personalization utility.
\end{abstract}

\section{Introduction}

Long-term memory is important for autonomous agentic systems, particularly as Large Language Model (LLM) agents are increasingly deployed for complex, multi-session, and long-horizon tasks~\cite{park2023generative,packer2023memgpt}. To enable agents with persistent capabilities across extended interactions, a variety of frameworks emerged, ranging from hierarchical key-value stores to dynamic context paging systems~\cite{wang2023augmenting,modarressi2023ret,zhong2024memorybank}. Concurrently, extensive benchmarking efforts scrutinized memory hallucinations~\cite{dziri2022origin}, factual retrieval precision~\cite{liu2023think}, and long-context recall degradation~\cite{liu2024lost,maharana2024evaluating}. Despite these advancements, agents deployed in open-ended lifelong interactions often exhibit memory behavior that remains frustratingly misaligned with user expectations~\cite{zhang2025survey}.

\begin{figure}[!t]
\centering
\includegraphics[width=\columnwidth]{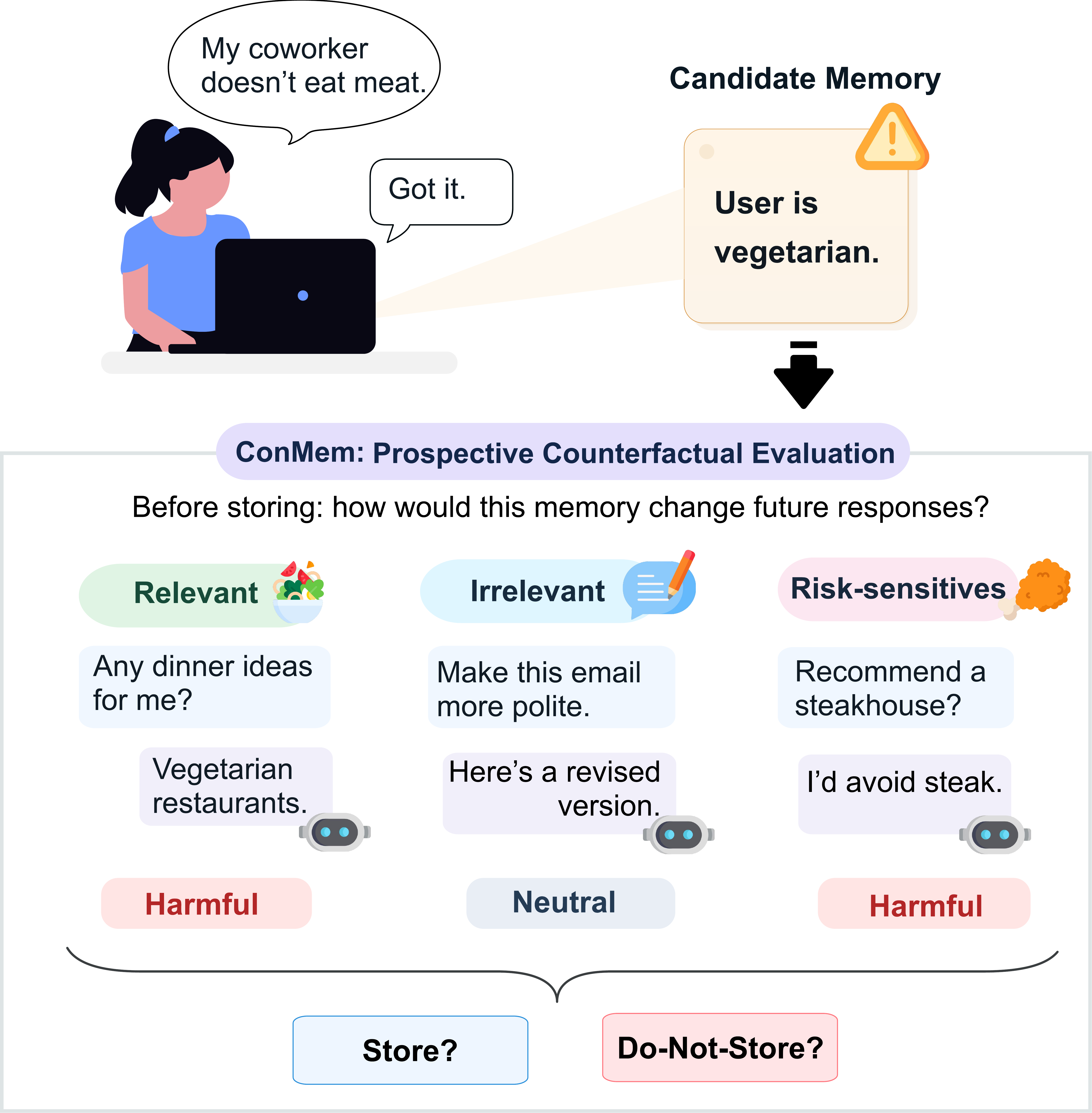}
\caption{An illustration of memory misalignment.}
\label{fig:teaser}
\vspace{-10pt}
\end{figure}

As seen in Figure~\ref{fig:teaser}, imagine a user casually conversing with a personal assistant about planning a farewell dinner for a vegetarian colleague: \textit{``can you suggest a fine-dining spot? My co-worker doesn't eat meat.''} A naive memory pipeline, eager to extract personal attributes, commits to long-term storage: \textit{``user is strictly vegetarian.''} In a subsequent session three weeks later, when the user requests recommendations of a steakhouse for their own birthday, the agent filters out all beef dishes, warns the user about dietary violations, and refuses to provide relevant suggestions. In this scenario, the stored information is a misattributed inference. When later recalled, such inappropriate memories pollute the agent's future decision boundaries.

We frame this challenge as the \textbf{memory misalignment} problem. Extending the taxonomy established by~\citet{chen2026characterizing}, memory misalignment spans multiple failure classes across two operational stages: \emph{memory generation} (e.g., attribution confusion, over-inference, and granularity mismatch) and \emph{memory usage} (e.g., context misapplication, stale memory reliance, and privacy overuse). Crucially, existing memory techniques evaluate candidate memories in isolation, relying primarily on extraction confidence, semantic relevance, or surface-level factual heuristics~\cite{salama2025meminsight,tan2025membench}. Therefore, they fail to anticipate how a candidate memory will affect downstream agent behavior across future interactions, leading either to over-retention or over-pruning.

To benchmark memory admission and usage within multi-turn interaction, we construct \textbf{MemAlign}. Building upon~\citet{chen2026characterizing}, we curate 1,079 candidate memories and augment each record with a multi-session conversational history generated via Qwen3.8-Max \citep{qwen38}. All conversational contexts pass structural checks, LLM-based semantic audits, and human review, which improve fidelity to the source annotations and reduce label-revealing shortcuts. These dialogues preserve the epistemic strength, entity attribution, temporal evolution, and corrective feedback of the source interaction while prohibiting the lexicalization of gold storage decisions.

To mitigate memory misalignment at the admission stage, we propose \textbf{ConMem}, a memory admission framework based on counterfactual probing of future interactions. Instead of treating admission as a static text-classification task, ConMem decides whether to store a candidate memory by simulating future interactions and comparing responses generated with and without access to the memory. These comparisons are aggregated into the Counterfactual Memory Treatment Effect (CMTE), a score that reflects the net downstream effect of memory access. Given a candidate memory and its historical context, ConMem synthesizes three future probes: (i) a relevant probe that tests whether the memory provides useful personalization, (ii) an irrelevant probe that checks whether the memory induces unnecessary personalization where it should not matter, and (iii) a risk-sensitive probe that introduces entity, temporal, or scope shifts to expose inappropriate memory carry-over. A blind quality judge and a separate attribution auditor then determine whether each response change is beneficial and whether it is attributable to the memory.

Extensive evaluations on MemAlign show that ConMem outperforms state-of-the-art memory mechanisms, including A-MAC~\cite{zhang2026adaptive} and SAGE~\cite{peng2026sagemultiagentselfevolutionllm}. Specifically, ConMem achieves an accuracy of 75.16\%, a balanced accuracy of 64.15\%, and a Macro-F1 of 65.31\%, surpassing A-MAC by 15.40 percentage points in Macro-F1 while effectively breaking the extreme retention biases of existing baselines. Furthermore, granular error analyses provide empirical insights into the delicate trade-off between memory utility and safety, showing robust generalization across parameter spaces while highlighting nuanced contextual scope boundaries as important directions for future exploration.

Collectively, this paper contributes: (i) we formalize the agent memory misalignment problem and introduce MemAlign, a benchmark of 1,079 candidate memories grounded in dual-validated multi-session dialogues; (ii) we propose ConMem, reframing memory admission as prospective counterfactual evaluation across complementary future scenarios; (iii) we show that \textsc{ConMem} substantially outperforms existing baselines by achieving 65.31\% Macro-F1, while uncovering behavioral trade-offs across distinct misalignment categories.

\section{Related Work}

\subsection{Memory of Agentic System}

Recent architectural advancements have established structured and hierarchical memory frameworks as core components for enabling dynamic context management in LLM agents. ~\citet{park2023generative} pioneered agentic memory design by simulating human-like behavior through social context interactions. To go beyond static prompts, MemGPT~\cite{packer2023memgpt} introduces an operating system-inspired virtual memory management system that enables agents to dynamically page and edit memory across storage tiers via function calling. Complementary architectures like LONGMEM~\cite{wang2023augmenting} leverage a decoupled trainable side network to retrieve long-term textual contexts, while RET-LLM~\cite{modarressi2023ret} equips models with an external, read-write memory unit via direct API interactions. Furthermore, TiM~\cite{liu2023think} improves reteieval efficiency by storing abstracted, structured ``thoughts'' rather than unstructured conversation history, and CoPS~\cite{zhou2024cognitive} applies a tripartite human-inspired memory system to organize search histories for personalized retrieval. 

Beyond structural storage, agents increasingly rely on experiential learning and domain-specific state tracking to evolve dynamically without weight updates. Reflexion~\cite{shinn2023reflexion} optimizes agent decision-making by storing natural language self-reflections of past failures in episodic memory. Similarly, ExpeL~\cite{zhao2024expel} extracts high-level insights from successful trajectory memories in vector stores to enable cross-task generalization. Incorporating cognitive mechanisms, MemoryBank~\cite{zhong2024memorybank} applies Ebbinghaus's forgetting curve to guide memory update and retention strategies. Expanding into physical and multimodal environments, JARVIS-1~\cite{wang2024jarvis} uses a multimodal memory component for lifelong planning in open-world tasks, MuRAG~\cite{chen2022murag} lays the ground for cross-modal retrieval-augmented reasoning, and~\citet{fan2024videoagent} proposes a unified memory mechanism tracking temporal segment descriptions and object states for video understanding.

To rigorously evaluate long-term contextual retainment and retrieval fidelity, several benchmarking suits have been developed. LOCOMO~\cite{maharana2024evaluating} assesses how agents reflect on and retrieve multi-session personalized information over extensive temporal horizons. Similarly, LongMemEval~\cite{wu2024longmemeval} probes the memory capacity of chat assistants and analyzes optimized indexing and reading strategies. Extending these evaluations across those criteria, MemBench~\cite{tan2025membench} benchmarks agentic memory mechanisms using multi-scenario, multi-level, and multi-metric dimensions. Addressing localized memory extraction, MemInsight~\cite{salama2025meminsight} evaluates autonomous memory augmentation by generating attribute-based representations.

Reviews provide formal taxonomies and conceptual baselines for categorizing diverse agentic memory mechanisms.~\citet{zhang2025survey} present a comprehensive review mapping out agentic memory implementations, evaluation methodologies, and current boundaries. Similarly,~\citet{wang2024survey} classify autonomous agent memory modules through the tri-fold lens of memory structures, formats, and core operations. From a holistic perspective,~\citet{xi2025rise} formulated a general agent framework comprising brain, perception, and action modules, situating long-term memory with societal behavior and prospective challenges.

\subsection{Alignment and Misalignment}

Retrieval-augmented mechanisms inherently suffer from contextual bottlenecks and factual hallucinations that undermine system reliability.~\citet{shuster2021retrieval} showed that grounding dialogue in neural retrieval mitigates hallucination, though~\citet{dziri2022origin} and~\citet{zhang2025siren} highlight persistent factual error patterns across the LLM lifecycle. In terms of structural context limitations,~\citet{liu2024lost} identify the ``lost in the middle'' phenomenon, showing a U-shaped performance curve where LLMs struggle to recall information positioned in the middle of long sequences. To evaluate these boundaries, RGB~\cite{chen2024benchmarking} introduces a benchmark targeting RAG capabilities in noise robustness and negative rejection.

Integrating persistent, updatable memory exposes agents to novel adversarial vectors, privacy leaks, and data manipulation attacks. AGENTPOISON~\cite{chen2024agentpoison} introduces a backdoor attack that pollutes an agent's memory to force malicious actions, while~\citet{hubinger2024sleeper} prove that backdoor behaviors can stubbornly survive standard safety fine-tuning. Addressing memory-specific threats,~\citet{sunil2026memory} evaluate memory injection attacks (MINJA) on healthcare agents and propose trust-aware memory sanitization. Tool-calling vulnerabilities are analyzed in InjecAgent~\cite{zhan2024injecagent}, Indirect Prompt Injection (IPI) threats introduced by~\citet{greshake2023not}, and stateful environmental security evaluated via AgentDojo~\cite{debenedetti2024agentdojo}. Furthermore, MEXTRA~\cite{wang2025unveiling} exposes privacy risks by demonstrating black-box extraction of sensitive conversation history from agent memory, complementing real-world empirical findings by~\citet{jo2024understanding} regarding memory-induced user privacy concerns in public health chatbots. 

Mitigating these safety risks requires robust alignment frameworks, structural defense, and safe design paradigms. 
InstructGPT~\cite{ouyang2022training} establishes the foundation for aligning LLMs with user intent via Reinforcement Learning from Human Feedback (RLHF). 
Addressing explicit instruction safety,~\citet{wallace2024instruction} propose an instruction hierarchy framework to enforce system instruction precedence over lower-privileged user or tool inputs, thereby defending against prompt injections. 
From a holistic architectural perspective,~\citet{li2024personal} review the intelligence levels, capabilities, and security boundaries of personal agents, while~\citet{amodei2016concrete} categorize foundational empirical machine learning safety risks to guide trustworthy system deployment.


\section{MemAlign}

\begin{figure}[!htbp]
    \centering
    \includegraphics[width=\columnwidth]{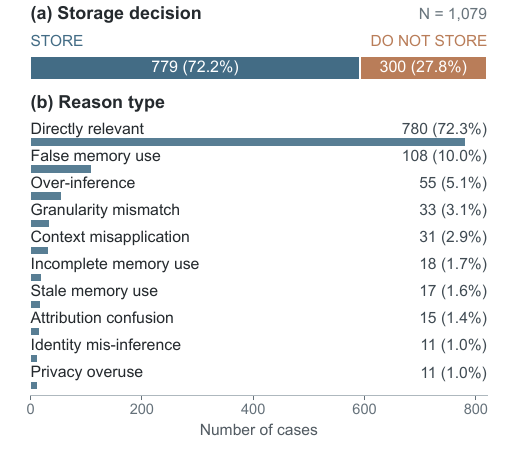}
    \caption{Distribution of (a) storage decisions and
    (b) reason types in \textsc{MemAlign} ($N=1{,}079$).
    Labels indicate instance counts and percentages.}
    \label{fig:dataset_distribution}
    \vspace{-10pt}
\end{figure}

We build \textbf{MemAlign} on the dataset introduced by~\citet{chen2026characterizing}. The task is a binary decision of whether a candidate memory should be stored for future personalization. Given a candidate memory and its historical interaction context, a model predicts \textsc{Store} or \textsc{Do Not Store}, with the original user-annotated storage decision serving as the reference label (see Figure~\ref{fig:dataset_distribution} for the label and reason-type distributions). Each original record contains a candidate memory, an accuracy annotation, a storage decision, and brief user feedback on the memory. To provide the conversational grounding absent in the source data, we synthesize a multi-turn historical interaction context for each instance while retaining all original annotations.


\vspace{0.1cm}
\noindent\textbf{Context Generation.}
We employ Qwen3.8-Max \citep{qwen38} to synthesize conversational histories conditioned on four pieces of \emph{construction supervision}: the candidate memory, user feedback, assigned reason type, and gold storage decision. These inputs serve to reconstruct the latent evidentiary relationship between the generated dialogue and the candidate memory, not to reveal the gold storage label to downstream admission models.

Specifically, the generated context must faithfully mirror the epistemic certainty, entity attribution, temporal status, and scope implied by the evidence. For instance, over-inferred memories are backed only by weak or indirect premises, attribution-confusion cases maintain strict entity ownership, stale memories incorporate explicit temporal shifts, and inaccurate memories retain the contradictions noted in user feedback.

To prevent label leakage and artificial artifacts, the model is strictly constrained from lexicalizing metadata, such as storage labels, reason categories, or retention preferences, echoing the candidate memory verbatim, amplifying recurrence or permanence, or fabricating personal details. The synthesized dialogue therefore features natural interactional cues, such as routine requests, corrections, and follow-ups, rather than summary-level persona profiles. Furthermore, generation prompts enforce session parsimony, where dialogues use one session by default, extending to two or three only when modelling temporal updates or independent interactions demands it. Session counts are explicitly decoupled from the assigned reason type, and the source memory's language is preserved.


\begin{figure*}[!htbp]
\centering
\includegraphics[width=0.98\textwidth]{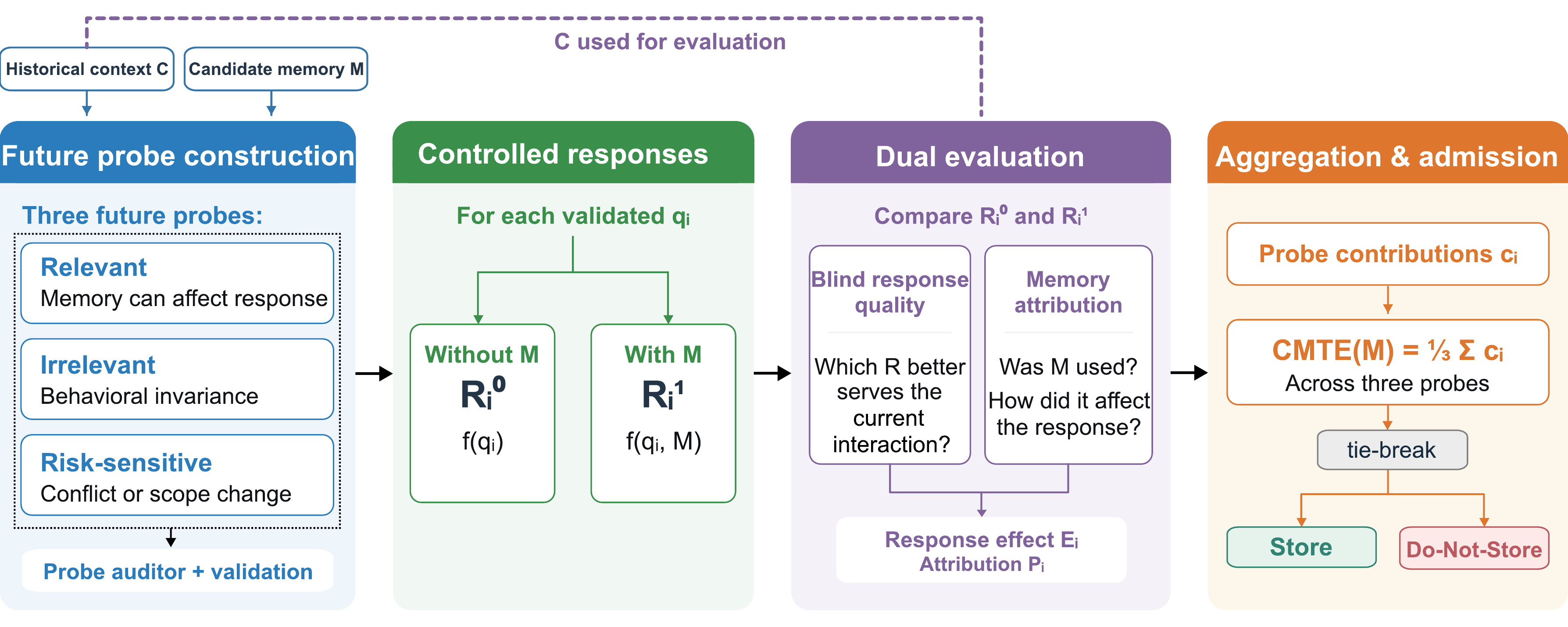}
\caption{Overview of ConMem. Given a historical context $C$ and a candidate memory $M$, ConMem (1) constructs and validates three future probes (relevant, irrelevant, and risk-sensitive); (2) generates a control response $R_i^0$ without $M$ and a treatment response $R_i^1$ with $M$ for each probe $q_i$; (3) compares the two responses with a blind quality judge and a memory-attribution auditor; and (4) aggregates the signed probe contributions $c_i$ into $\mathrm{CMTE}(M)$ to decide \textsc{Store} or \textsc{Do-Not-Store}, invoking a tie-break judge when the score is zero. $C$ is withheld from response generation and used only for evaluation.}
\label{fig:pipeline}
\vspace{-10pt}
\end{figure*}

\vspace{0.1cm}
\noindent\textbf{Validation.}
To assess and improve context quality, we combine two-stage
automated validation with human review. Structural checks
require one to three sequentially indexed sessions, each
containing two, four, or six non-empty messages with alternating
user and assistant roles. Each session starts with a user
utterance and ends with an assistant response, and the total
context contains at most 12 messages. Additional checks screen
for near-verbatim reproduction of the candidate memory and
explicit benchmark taxonomy leakage.

We then conduct semantic validation using GPT-6 Astra.
The audit assesses eight dimensions: evidence fidelity,
reason-type consistency, candidate-memory mirroring,
diagnostic shortcuts, annotation lexicalization, retention
leakage, reason-specific templating, and conversational
naturalness and conciseness. Acceptance requires fidelity
and consistency with the available evidence, natural dialogue,
and the absence of the identified artifacts.
Contexts requiring revision are refined using structural
error reports and semantic feedback. Across the 1,081 initial
instances, generation involved one, two, three, or four
attempts for 898, 155, 16, and 12 instances, respectively.
These counts include two instances that received manual
supplements and were subsequently excluded.

Post-generation refinement included restoring richer
reference contexts for 139 instances and manually rewriting
seven additional contexts to address unnatural compression,
artificial diagnostic cues, or unsupported details.
Two researchers independently reviewed the generated contexts
against the same quality criteria, resolving disagreements
through discussion. The review concerned the synthesized
contexts; the original accuracy and storage annotations
remained unchanged. The two manually supplemented instances
were excluded because the source feedback provided
insufficient evidence to ground their contexts, leaving
1,079 instances in the final benchmark.
The final dataset contains 779 \textsc{Store} instances
(72.2\%) and 300 \textsc{Do-Not-Store} instances (27.8\%),
spanning ten reason categories.
Figure~\ref{fig:dataset_distribution} presents storage decisions' and reason types' distributions.

\section{ConMem}
\label{sec:cmte}

\subsection{Overview}

ConMem determines whether a candidate memory should be retained by estimating its \textbf{Counterfactual Memory Treatment Effect (CMTE)} on future interactions. Given historical context $C$ and candidate memory $M$, ConMem constructs controlled future scenarios and compares responses generated with and without access to $M$. Rather than predicting storage suitability directly from the memory text, it evaluates whether access to the memory produces useful, unnecessary, or risky behavioral changes.

The framework consists of four stages, as illustrated in Figure~\ref{fig:pipeline}: future probe construction, controlled response generation, dual evaluation of response quality and memory attribution, and counterfactual aggregation for admission. Detailed prompts are provided in Appendix~\ref{app:prompts}.

\subsection{Future Probe Construction}

Because memory utility is inherently context-dependent, each candidate memory is evaluated across a probe triplet $Q=\{q_{\mathrm{rel}},q_{\mathrm{irr}},q_{\mathrm{risk}}\}$. The \textit{relevant} probe creates a realistic setting in which $M$ provides non-redundant user-specific information that materially affect the response. The \textit{irrelevant} probe represents a task whose response quality should not depend on $M$. The \textit{risk} probe introduces an observable change (e.g., target entity shifts, overrides, or scope transitions) designed to expose inappropriate memory carry-over. 

\vspace{0.1cm}
\noindent\textbf{Probe Validation.}
An independent probe auditor verifies that generated queries fulfill their designated semantic roles while ensuring no leakage of candidate memories, gold storage labels, or benchmark artifacts.
Invalid probes are revised before evaluation. We first regenerate the probe set while preserving valid probes, then repair individual invalid roles when necessary, and use a fresh reconstruction only for cases that remain invalid. 
This guarantees downstream treatment effects reflect memory utility rather than flawed probes.

\subsection{Controlled Counterfactual Responses}

To isolate the marginal causal effect of candidate memory $M$, control and treatment generations differ exclusively by memory access: $R_i^{0}=f(q_i)$, $R_i^{1}=f(q_i,M)$. Both branches use identical model configurations. The treatment condition receives $M$ as optional context alongside instructions to apply it solely when relevant and appropriate. Historical context $C$ is withheld from both generation branches to isolate the direct causal impact of $M$, and is reserved for downstream evaluation to assess whether memory-induced personalization is substantiated by prior interactions.



\subsection{Dual Evaluation of Memory Effects}

As generation variance can arise independently of memory retrieval, ConMem decouples response quality assessment from memory attribution.

\vspace{0.1cm}
\noindent\textbf{Blind Response Quality.}
A blind pairwise judge evaluates $R_i^{0}$ and $R_i^{1}$ against historical context $C$ and probe $q$. The judge is agnostic to $M$ and treatment-assignment conditions. Generations are presented in deterministically randomized order (positions A/B) to counter position bias, yielding a preference of \textbf{A}, \textbf{B} or \textbf{Tie}.

\vspace{0.1cm}
\noindent\textbf{Memory Attribution.}
An attribution auditor evaluates $C$, $M$, $q_i$, $R_i^{0}$, and $R_i^{1}$ to ascertain if $M$ was substantively invoked and materially altered outputs, classifying behavioral effects into: $P_i \in \{\text{utility}, \text{unnecessary}, \text{neutral}, \text{risk}\}$
where \textit{Utility} signifies accurate, appropriately bounded personalization; \textit{unnecessary} denotes benign yet irrelevant incorporation, \textit{risk} captures negative externalities such as entity confusion or over-generalization,  and \textit{neutral} indicates no material intervention.

Decoupling these two judgments prevents generation quality variation from being interpreted as evidence of memory misalignment. We define the response-level effect $
E_i\in
\{\textsc{Helpful},\textsc{Neutral},\textsc{Harmful}\}.
$ If the memory is not substantively used or does not change the treatment response, $E_i$ is \textsc{Neutral}. Otherwise, it is \textsc{Helpful} when the judge prefers treatment response, \textsc{Harmful} when preferring control response, and \textsc{Neutral} for a tie.

\subsection{Counterfactual Scoring and Admission}

\noindent\textbf{Probe Contributions.}
Each probe contributes a signed effect according to its semantic role, attributed use, and observed response quality. For a relevant probe, helpful utility contributes $+1.0$, while attributed risk contributes $-1.0$. For an irrelevant probe, unnecessary personalization contributes $-0.5$ and attributed risk contributes $-1.0$. For a risk probe, helpful utility contributes $+0.5$, unnecessary personalization contributes $-0.5$, and attributed risk contributes $-1.5$. All other outcomes contribute $0$.
Positive evidence therefore requires both causal attribution of memory use and response quality improvement. Negative evidence requires explicit attribution of unnecessary or risky use. The aggregate metric is computed as: $\mathrm{CMTE}(M) = \frac{1}{K}\sum_{i=1}^{K}c_i, \qquad K=3.$

\paragraph{Score-First Admission.}
A positive ConMem score yields \textsc{Store}, while a negative score yields \textsc{Do-Not-Store}:
When $|\mathrm{CMTE}(M)|\leq\epsilon$, evidence is considered inconclusive. ConMem invokes a parameter-independent tie-break judge that evaluates C, M, and raw counterfactual logs. This tie-breaker is shielded from predetermined numeric weights and gold labels, preventing heuristic bias.

\vspace{0.1cm}
\noindent\textbf{Reason Classification.}
Following admission, ConMem executes a post-hoc classification to assign a structured explanation, but does not modify $\hat{y}$ or $\text{CMTE}(M)$. Accepted cases (\textsc{Store}) map to \texttt{directly\_\allowbreak relevant}. For rejected cases (\textsc{Do-Not-Store}), a classifier takes $C$, $M$, $\hat{y}$, and the generation rationale to assign the primary failure category: \texttt{context\_\allowbreak misapplication}, \texttt{memory\_\allowbreak omission}, \texttt{granularity\_\allowbreak mismatch}, \texttt{over\_\allowbreak inference}, \texttt{stale\_\allowbreak memory\_\allowbreak use}, \texttt{privacy\_\allowbreak refusal}, \texttt{privacy\_\allowbreak overuse}, \texttt{attribution\_\allowbreak confusion}, \texttt{identity\_\allowbreak misinference}, \texttt{incomplete\_\allowbreak memory\_\allowbreak use}, or \texttt{false\_\allowbreak memory\_\allowbreak use}. These categorical attributions are primarily used for downstream interpretation and failure analysis.

\section{Experiments}

\subsection{Methods}

We compare the following methods:

$\bullet$ ConMem: as introduced in Sec.~\ref{sec:cmte}.

$\bullet$ SAGE: a multi-agent reasoning framework (Challenger, Planner, Solver, Critic) tested here for its deliberation capability in memory admission~\cite{peng2026sagemultiagentselfevolutionllm}.

$\bullet$ A-MAC: an admission control method scoring candidates by expected utility, factual confidence, semantic novelty, recency, and content-type priors~\cite{zhang2026adaptive}.

$\bullet$ A-MEM~\cite{xu2025mem}: organizes dialogue into attributed memories. Following~\citet{zhang2026adaptive}, candidates are scored via normalized importance and semantic novelty.

$\bullet$ MemoryBank: ~\citet{10.1609/aaai.v38i17.29946} models retention via recency, relevance, and importance. It followed~\citet{zhang2026adaptive} by scoring candidates via semantic relevance and estimated utility.

\subsection{Parameters and Settings}

All ConMem components and baseline LLM backbones use Qwen3.8-Max with temperature 0, seed 42, and default API decoding parameters. Output token limits are capped at 2,048 for probing, validation, auditing, and tie-breaking stages, and 8,192 for control and treatment response generation. We employ medium reasoning effort for probe, response, and quality generation, and high effort for validation, auditing, and tie-breaking. Each candidate is evaluated across three scenario probes with a zero-score tolerance of $\epsilon=10^{-12}$, requiring 14 standard model calls plus an optional tie-break call, with detailed efficiency metrics appear in Appendix~\ref{app:efficiency}. For baselines evaluated on the same 1,079 candidates, A-MAC~\cite{zhang2026adaptive} adopts its published five-feature policy with weights (0.1, 0.1, 0.1, 0.1, 0.6), and a 0.55 threshold. A-MEM~\cite{xu2025mem} computes admission scores as 0.7$I_{norm}$+0.3$N$ (threshold 0.5) from extracted attributes. MemoryBank~\cite{zhong2024memorybank} uses 0.3$R$ + 0.4$L$ + 0.3$I$, adapting recency to benchmark constraints. SAGE~\cite{peng2026sagemultiagentselfevolutionllm} evaluates decisions from its multi-agent reasoning.

\subsection{Results}

\subsubsection{Overall Performance}

Table~\ref{tab:main_results} compares ConMem with MemoryBank, A-MEM, A-MAC, and SAGE on the 1,079 candidate memories. ConMem achieves 75.16\% accuracy, 64.15\% balanced accuracy, and 65.13\% Macro-F1, outperforming all baselines on accuracy and Macro-F1. Relative to the strongest accuracy baseline, A-MAC, ConMem yields gains of 5.83, 12.23 and 15.40 percentage points across accuracy, balanced accuracy and Macro-F1, respectively. Compared to MemoryBank, the leading baseline in balanced accuracy, ConMem improves accuracy by 3.97 points and Macro-F1 by 1.01 points.

Class-wise performance highlights contrasting admission behaviors. A-Mem and A-MAC exhibit an over-retention bias, attaining high \textbf{Store} recall (88.83\% and 91.12\%) at the expense of \textbf{Do-Not-Store} recall (11.67\% and 12.71\%). Conversely, MemoryBank offers a more balanced trade-off (69.67\% \textbf{Do-Not-Store} and 66.24\% \textbf{Store} recall), while SAGE excessively rejects candidates, capturing only 34.15\% of valid \textbf{Store} memories. ConMem achieves the best overall trade-off, sustaining a high \textbf{Store} recall of 88.96\% while increasing \textbf{Do-Not-Store} recall to 39.33\%, thereby securing the highest balanced accuracy and Macro-F1.

\begin{table}[!htbp]
\centering
\caption{Performance comparison across techniques.}
\label{tab:main_results}
\scriptsize
\begin{tabular}{lccccc}
\toprule
Method & Acc. & BAcc. & Macro-F1 & Store R. & DNS R. \\
\midrule

MemoryBank
& 67.19 & 67.95 & 64.30 & 66.24 & \textbf{69.67} \\

A-MEM
& 67.38 & 50.25 & 48.16 & 88.83 & 11.67 \\

A-MAC
& 69.33 & 51.91 & 49.91 & \textbf{91.12} & 12.71 \\

SAGE
& 40.96 & 46.41 & 40.55 & 34.15 & 58.67 \\

ConMem
& \textbf{75.16} & 64.15 & \textbf{65.31} & 88.96 & 39.33 \\

\bottomrule
\end{tabular}
\end{table}

\subsubsection{Cross-Dataset Generalization}

\paragraph{LoCoMo.}
We evaluated all methods on LoCoMo, where ConMem achieves the highest precision of 41.8\%, outperforming competitors such as A-MAC (41.7\%) and A-MEM (37.1\%). While methods like A-MEM and A-MAC yield higher recall (100.0\% and 97.2\%, respectively), ConMem maintains a balanced trade-off with 75.7\% recall and a 53.9\% F1 score. This substantially outperforms retrieval baselines like MemGPT (32.4\% F1) and MemoryBank (45.2\% F1), rivaling dedicated frameworks and confirming its strong generalizability across memory management benchmarks.



\paragraph{HaluMem.}
We further evaluate ConMem on HaluMem, which separates valid reference memories from interfering ones. On a randomly selected balanced 400-case subset, ConMem achieves 66.00\% accuracy, 66.00\% balanced accuracy, and 63.22\% Macro-F1, retaining 93.50\% of valid memories while rejecting 38.50\% of distractor memories. These results show ConMem's efficacy under distinct criteria targeting memory interference, while the lower \textbf{Do-Not-Store} recall reflects an ongoing inclination toward over-retention.

\subsubsection{Error Analysis}
ConMem achieves 881/1,079 accuracy, with errors heavily skewed toward over-retention: 182 false positives (60.67\% of \textit{Do-Not-Store}) versus 86 false negatives (11.04\% of \textit{Store}). This disparity reflects asymmetric counterfactual scoring. False Positives stem from unpenalized utility, where strong relevance signals (mean +0.852) overshadow undetected risks (risk mean +0.107). Conversely, FNs result from disproportionate risk penalties (mean -1.052), which override valid relevance signals (mean +0.360) and lead to erroneous discards.

Rejection efficacy varies substantially across error types (Table~\ref{tab:error_types}). ConMem effectively flags factual inconsistencies (e.g., \textit{state\_memory\_use} at 58.82\% recall and \textit{false\_memory\_use} at 57.41\%), but underperforms on fine-grained contextual and scoping boundaries, including \textit{context\_misapplication} (12.90\%) and \textit{incomplete\_memory\_use} (16.67\%). Notably, tie-breaking is not the primary error source, where the admission judge achieves 81.91\% accuracy on the 199 tie-break cases. Misclassifications primarily stem from upstream counterfactual evidence generation and aggregation than final arbitration.

\begin{table}[!htbp]
\centering
\small
\caption{ConMem performance across rejection categories.}
\label{tab:error_types}
\begin{tabular}{lrr}
\toprule
Reason type & $N$ & Recall (\%) \\
\midrule
context\_misapplication  & 31  & 12.90 \\
incomplete\_memory\_use  & 18  & 16.67 \\
granularity\_mismatch    & 33  & 21.21 \\
identity\_mis\_inference & 11  & 27.27 \\
privacy\_overuse         & 11  & 27.27 \\
attribution\_confusion   & 15  & 33.33 \\
over\_inference          & 55  & 38.18 \\
false\_memory\_use       & 108 & 57.41 \\
stale\_memory\_use       & 17  & 58.82 \\
\bottomrule
\end{tabular}
\end{table}


\subsubsection{Parameter Sensitivity}

We evaluate ConMem over 27 pre-specified combinations of its three scoring parameters, with the setting fixed before evaluation. Across these settings, accuracy spans 75.16\%--78.13\%, balanced accuracy 64.15\%--65.69\%, and Macro-F1 65.31\%--67.50\%. \textsc{Store} recall varies from 88.96\% to 93.71\%, while \textsc{Do-Not-Store} recall remains between 37.33\% and 39.67\%. Overall, ConMem exhibits strong parameter stability, with the default configuration achieving 75.16\% accuracy, 64.15\% balanced accuracy, and 65.31\% Macro-F1. Complete results are detailed in Appendix~\ref{app:sensitivity}.

\subsubsection{Backbone Robustness}

We evaluate ConMem with three backbones, Qwen3.8-Max, GPT-5.6-sol, and DeepSeek-V4-Flash, on the same 1,079 cases under identical prompts, evaluation protocols, scoring weights, and admission rules. GPT-5.6-sol achieves the highest accuracy (77.85\%), balanced accuracy (70.93\%), and Macro-F1 (71.54\%), followed by DeepSeek-V4-Flash (76.83\%, 67.76\%, and 68.89\%) and Qwen3.8-Max (75.16\%, 64.15\%, and 65.31\%). For DeepSeek, sensitivity analyses over 27 predefined cost configurations yields stable accuracies (76.65\%--77.94\%), confirming robustness to coefficient variations. Full results appear in Appendix~\ref{app:backbone}.

\subsubsection{Ablation Study}
To assess ConMem's validation and decision modules, we evaluate four variants on the 1,079 cases using original probe sets and weights: (1) \textit{w/o Quality Judge}, which directly uses the attribution auditor's utility assessment without blind quality confirmation; (2) \textit{w/o Attribution Auditor}, which derives utility scores solely from pairwise preferences without causal attribution verification; and (3--4) \textit{Zero $\rightarrow$ STORE} and \textit{Zero $\rightarrow$ DO\_NOT\_STORE}, replacing zero-score tie-breaker with fixed defaults.

\begin{table}[!htbp]
\centering
\caption{Ablation results for ConMem's validation and decision components (\%).}
\label{tab:cmte_ablation}
\small
\begin{tabular}{lccc}
\toprule
\textbf{Variant} & \textbf{Acc.} & \textbf{BAcc.} & \textbf{Macro-F1} \\
\midrule
Full ConMem & \textbf{75.16} & \textbf{64.15} & \textbf{65.31} \\
w/o Quality Judge & 72.57 & 59.28 & 59.87 \\
w/o Attribution Auditor & 55.05 & 55.55 & 52.41 \\
Zero $\rightarrow$ STORE & 71.73 & 56.85 & 56.83 \\
Zero $\rightarrow$ DO\_NOT\_STORE & 66.82 & 60.93 & 60.38 \\
\bottomrule
\end{tabular}
\end{table}

As shown in Table~\ref{tab:cmte_ablation}, removing the quality judge reduces Macro-F1 from 65.31\% to 59.87\%, while removing the attribution auditor degrades it further to 52.41\%, showing the synergy between quality confirmation and causal attribution. Moreover, replacing adaptive tie-breaking with fixed decisions degrades performance, where a default SCORE inflates false-positive retention, and a default DO\_NOT\_STORE leads to under-admission. These results highlight the need to couple evidence verification with adaptive thresholding.

\section{Conclusion}
This paper tackles memory misalignment in LLM chatbots, where conventional memory admission fails to discern whether candidate memories are appropriate. To address this, we introduce \textbf{MemAlign}, a benchmark of 1,079 memories grounded in verified multi-session dialogues, alongside \textbf{ConMem}. grounded in counterfactual memory treatment effects, \textbf{ConMem} evaluates candidates by simulating causal interventions across diverse future probes spanning utility, invariance, and risk, while decoupling blind quality assessment from attribution auditing. Empirical results show that \textbf{ConMem} consistently outperforms competitive baselines, achieving 75.16\% accuracy, 64.15\% balanced accuracy, and 65.31\% Macro-F1.

\clearpage
\section*{Limitations}

We acknowledge several limitations of this paper. First, although verified, the memory context is synthetic, which may differ from realistic chat histories. We made this choice because users can hardly memorize their histories, and those histories also pose privacy concerns. Future work could build upon our work to build such traces.
Second, we primarily target the memories for LLM-based chatbots that  face Chinese users, noting that memories in other agentic systems may be different and warrant future examination.

\section*{Ethics Statement}

This work complies with the ACL Ethics Policy. The multi-session conversational histories constructed for MemAlign are entirely synthetic and contain no personally identifiable information (PII). All user attributes, preference profiles, and interaction trajectories represent simulated scenarios designed strictly for probing memory misalignment taxonomies. No private personal logs, proprietary user communications, or real-world sensitive records were collected, crawled, or utilized during benchmark construction or evaluation.

From a broad societal perspective, deploying persistent memory mechanisms in autonomous agents introduces systemic risks regarding privacy leakage, unwarranted personalization, and the perpetuation of misattributed user traits. Rather than exacerbating these hazards, ConMem is intentionally architected as an alignment and safety mechanism. By enforcing behavioral invariance across irrelevant inquiries and explicitly penalizing inappropriate memory carry-over under contextual boundary shifts, the framework actively prevents privacy overuse and over-retention, fostering trustworthy and boundary-aware conversational agent deployments.

\bibliography{custom}

\clearpage
\appendix

\twocolumn[
\begin{@twocolumnfalse}

\section*{Appendix}

\section{Parameter Sensitivity Details}
\label{app:sensitivity}


\begin{center}
\small
\captionof{table}{
Sensitivity results for the nine distinct combinations of the risk-safe
reward and explicit-risk penalty. Results are identical for
irrelevant-personalization costs of $0.25$, $0.50$, and $0.75$ when the
other parameters are fixed.
}
\label{tab:sensitivity_grid}

\setlength{\tabcolsep}{7pt}
\renewcommand{\arraystretch}{1.05}

\begin{tabular}{ccccccc}
\toprule
Safe Reward & Risk Penalty & Acc. & BAcc. & Macro-F1 & Store R. & DNS R. \\
\midrule
0.00 & 1.0 & 78.13 & 65.69 & 67.50 & 93.71 & 37.67 \\
0.00 & 1.5 & 75.25 & 64.31 & 65.49 & 88.96 & 39.67 \\
0.00 & 2.0 & 75.25 & 64.31 & 65.49 & 88.96 & 39.67 \\
\midrule
0.25 & 1.0 & 78.04 & 65.52 & 67.31 & 93.71 & 37.33 \\
0.25 & 1.5 & 75.16 & 64.15 & 65.31 & 88.96 & 39.33 \\
0.25 & 2.0 & 75.16 & 64.15 & 65.31 & 88.96 & 39.33 \\
\midrule
0.50 & 1.0 & 78.04 & 65.52 & 67.31 & 93.71 & 37.33 \\
0.50 & 1.5 & \textbf{75.16} & \textbf{64.15} & \textbf{65.31}
            & \textbf{88.96} & \textbf{39.33} \\
0.50 & 2.0 & 75.16 & 64.15 & 65.31 & 88.96 & 39.33 \\
\bottomrule
\end{tabular}
\end{center}

\vspace{0.8em}


\begin{center}
\small
\captionof{table}{
Overall sensitivity of ConMem to the scoring parameters.
}
\label{tab:sensitivity}

\begin{tabular}{lcc}
\toprule
Metric & Main setting & Range \\
\midrule
Accuracy              & 75.16 & 75.16--78.13 \\
Balanced Accuracy     & 64.15 & 64.15--65.69 \\
Macro-F1              & 65.31 & 65.31--67.50 \\
\textsc{Store} Recall & 88.96 & 88.96--93.71 \\
\textsc{DNS} Recall   & 39.33 & 37.33--39.67 \\
Zero-score Rate       & 18.44 & 18.44--25.58 \\
\bottomrule
\end{tabular}
\end{center}

\vspace{1em}

\end{@twocolumnfalse}
]

We perform a sensitivity analysis over 27 pre-specified combinations of three scoring parameters: the cost of unnecessary personalization on an irrelevant probe, the reward for safe and helpful memory use on a risk-sensitive probe, and the penalty for explicit risk on a risk-sensitive probe. Their tested values are $\{0.25,0.50,0.75\}$, $\{0,0.25,0.50\}$, and $\{1.0,1.5,2.0\}$, respectively. The main configuration uses $0.50$, $0.50$, and $1.50$. All settings are evaluated using the same counterfactual evidence, and the parameter-independent tie-break procedure is applied whenever the resulting ConMem score is zero.

The irrelevant-personalization cost has no effect on the final admission predictions within the tested range: changing it from $0.25$ to $0.75$ produces identical classification metrics when the other two parameters are fixed. The reward for safe memory use has only a small effect. Increasing it from $0$ to $0.25$ slightly changes the predictions, whereas increasing it further from $0.25$ to $0.50$ leaves the final classification metrics unchanged.

The largest variation is caused by the explicit-risk penalty on the risk-sensitive probe. With the other parameters fixed, reducing this penalty from $1.5$ to $1.0$ increases \textsc{Store} recall from $88.96\%$ to $93.71\%$, while \textsc{Do-Not-Store} recall decreases from approximately $39$--$40\%$ to $37$--$38\%$. Consequently, overall accuracy increases to approximately $78\%$, largely because \textsc{Store} cases are more frequent in the primary benchmark. Increasing the penalty from $1.5$ to $2.0$, in contrast, does not alter the final admission predictions. These results indicate that the observed variation primarily reflects the expected trade-off between preserving potentially useful memories and penalizing future misuse than broad instability of the scoring scheme.

Because the irrelevant-personalization cost does not affect the final predictions, Table~\ref{tab:sensitivity_grid} collapses the full 27-setting grid over this dimension and reports the nine distinct combinations of the remaining two parameters.

\section{Prompt Design}
\label{app:prompts}

This appendix summarizes the prompts used by ConMem. The prompts follow the functional modules described in Sec.~\ref{sec:cmte}. Benchmark labels are never provided to the probe generator, response agents, quality judge, or attribution auditor. Historical context, candidate memories, generated queries, and model responses are explicitly delimited and treated as data rather than instructions.

\subsection{Future Probe Generation}
\label{app:probe_prompt}

The probe generator receives the historical context and candidate memory and produces exactly one \textit{relevant}, one \textit{irrelevant}, and one \textit{risk} probe. The candidate memory is not assumed to be correct, stable, or suitable for storage.

For the relevant probe, the prompt asks for a realistic future query in which the candidate memory provides non-redundant user-specific information that could materially affect the response. The query itself must not reveal or paraphrase that information. For the irrelevant probe, the substantive task outcome should remain independent of the candidate memory. For the risk probe, the query must contain an observable condition---such as a different person, an updated fact, an explicit override, or a scope change---under which inappropriate carry-over can be detected. Hypothetical staleness alone is not sufficient.

The prompt further requires all probes to be natural user requests and prohibits references to benchmarks, storage decisions, treatment/control conditions, or hidden memory information.

\begin{Prompt}
Historical Context:
<<<HISTORICAL_CONTEXT>>>
{historical_context}
<<<END_HISTORICAL_CONTEXT>>>

Candidate Memory:
<<<CANDIDATE_MEMORY>>>
{candidate_memory}
<<<END_CANDIDATE_MEMORY>>>

Generate exactly three plausible future user queries:
one relevant, one irrelevant, and one risk probe.
\end{Prompt}

\subsection{Probe Validation and Repair}
\label{app:probe_validation_prompt}

A separate probe auditor checks the generated queries without making an admission decision. It verifies whether the relevant probe provides a non-redundant opportunity for memory use, whether the irrelevant probe is substantively memory-independent, whether the risk probe contains an observable conflict or scope change, and whether the probe set is free of memory or benchmark leakage.

When validation fails, the semantic audit is returned to the generator. Already valid probes are preserved when possible, while invalid roles are regenerated. If full-set regeneration remains unsuccessful, ConMem repairs the failed probe individually.

\begin{Prompt}
Historical Context:
<<<HISTORICAL_CONTEXT>>>
{historical_context}
<<<END_HISTORICAL_CONTEXT>>>

Candidate Memory:
<<<CANDIDATE_MEMORY>>>
{candidate_memory}
<<<END_CANDIDATE_MEMORY>>>

Role to repair:
{probe_type}

Current probe set:
<<<CURRENT_PROBES>>>
{current_probes}
<<<END_CURRENT_PROBES>>>

Semantic audit:
<<<SEMANTIC_AUDIT>>>
{semantic_audit}
<<<END_SEMANTIC_AUDIT>>>

Generate a replacement only for the requested role.
The replacement should correct the semantic failure
rather than merely paraphrase the previous query.
\end{Prompt}

If targeted repair still fails, the generator reconstructs a fresh three-probe set using the same role definitions.

\subsection{Counterfactual Response Generation}
\label{app:response_prompt}

The control and treatment branches use the same response instruction. The assistant is asked to answer the current request naturally and accurately, to use optional user-specific information only when relevant, and to prioritize the current request over historical information.

The control branch receives only the future query:

\begin{Prompt}
Current User Query:
<<<CURRENT_QUERY>>>
{query}
<<<END_CURRENT_QUERY>>>

Answer the current request.
\end{Prompt}

The treatment branch additionally receives the candidate memory:

\begin{Prompt}
Optional User-Specific Information:
<<<USER_INFORMATION>>>
{candidate_memory}
<<<END_USER_INFORMATION>>>

Current User Query:
<<<CURRENT_QUERY>>>
{query}
<<<END_CURRENT_QUERY>>>

Answer the current request.
\end{Prompt}

Thus, the two response-generation conditions differ only in whether the candidate memory is available.

\subsection{Blind Response-Quality Evaluation}
\label{app:quality_prompt}

The blind quality judge receives the historical context, future query, and two responses, but does not observe the candidate memory or know which response corresponds to the treatment condition. It evaluates correctness, relevance, appropriateness, and helpfulness, while treating the current query as the strongest evidence of present intent and constraints.

\begin{Prompt}
Historical Context:
<<<HISTORICAL_CONTEXT>>>
{historical_context}
<<<END_HISTORICAL_CONTEXT>>>

Future User Query:
<<<CURRENT_QUERY>>>
{query}
<<<END_CURRENT_QUERY>>>

Response A:
<<<RESPONSE_A>>>
{response_a}
<<<END_RESPONSE_A>>>

Response B:
<<<RESPONSE_B>>>
{response_b}
<<<END_RESPONSE_B>>>

Which response better serves this future interaction?
Return A, B, or TIE with a concise explanation.
\end{Prompt}

\subsection{Memory-Attribution Audit}
\label{app:attribution_prompt}

The attribution auditor observes the historical context, candidate memory, future query, control response, and treatment response. It does not compare response quality. Instead, it determines whether the candidate memory was substantively used, whether it materially changed the treatment response, and whether the resulting personalization is \textit{utility}, \textit{unnecessary}, \textit{neutral}, or \textit{risk}.

The auditor labels a change as \textit{utility} when it is relevant, supported by the historical context, and appropriately scoped. \textit{Unnecessary} denotes mild but irrelevant personalization. \textit{Risk} covers unsupported, misattributed, over-generalized, privacy-invasive, or conflicting memory use. If the memory is not materially used or does not change the response, the effect is \textit{neutral}.

\begin{Prompt}
Historical Context:
<<<HISTORICAL_CONTEXT>>>
{historical_context}
<<<END_HISTORICAL_CONTEXT>>>

Candidate Memory:
<<<CANDIDATE_MEMORY>>>
{candidate_memory}
<<<END_CANDIDATE_MEMORY>>>

Future User Query:
<<<CURRENT_QUERY>>>
{query}
<<<END_CURRENT_QUERY>>>

Control Response:
<<<CONTROL_RESPONSE>>>
{control_response}
<<<END_CONTROL_RESPONSE>>>

Treatment Response:
<<<TREATMENT_RESPONSE>>>
{treatment_response}
<<<END_TREATMENT_RESPONSE>>>

Determine whether the Candidate Memory was used,
whether it materially changed the treatment response,
and whether the resulting personalization is
utility, unnecessary, neutral, or risk.
\end{Prompt}

The auditor is explicitly instructed not to infer staleness merely because information could have changed. A risk judgment requires observable contradictory, updated, superseding, or otherwise inappropriate use.

\subsection{Zero-Score Admission Tie-Break}
\label{app:tiebreak_prompt}

The tie-break judge is invoked only when the aggregate CMTE score is zero. It receives the candidate memory, historical context, and raw counterfactual evidence, but does not receive the hand-specified CMTE weights or benchmark labels.

\begin{Prompt}
Candidate Memory:
<<<CANDIDATE_MEMORY>>>
{candidate_memory}
<<<END_CANDIDATE_MEMORY>>>

Historical Context:
<<<HISTORICAL_CONTEXT>>>
{historical_context}
<<<END_HISTORICAL_CONTEXT>>>

Raw Counterfactual Evidence:
<<<COUNTERFACTUAL_EVIDENCE>>>
{counterfactual_evidence}
<<<END_COUNTERFACTUAL_EVIDENCE>>>

The aggregate CMTE score is exactly zero and is therefore inconclusive. Make the STORE or DO_NOT_STORE judgment from the evidence above only.
\end{Prompt}

The judge favors \textsc{Store} when the candidate is sufficiently supported, appropriately attributed, and potentially reusable. It favors \textsc{Do-Not-Store} when the candidate is unsupported, misattributed, materially over-generalized, contradicted by context, or otherwise inappropriate to retain.

\subsection{Reason Classification}
\label{app:reason_prompt}

After the admission outcome is fixed, ConMem assigns a structured reason type for interpretation and error analysis. This stage is post-hoc and does not modify either the CMTE score or the admission prediction.

For a \textsc{Store} decision, the reason type is \texttt{directly\_relevant}. For a \textsc{Do-Not-Store} decision, the classifier receives the candidate memory, historical context, finalized admission prediction, and admission explanation:

\begin{Prompt}
You classify why this memory should NOT be stored.

Candidate Memory:
{candidate_memory}

Historical Context:
{historical_context}

CMTE Admission Prediction:
{prediction}

Admission Explanation:
{admission_reason}

Return:
- reason_type
- confidence
- a short explanation

Do not change the admission decision.
\end{Prompt}

The allowed rejection reasons are
\texttt{context\_misapplication},
\texttt{memory\_omission},
\texttt{granularity\_mismatch},
\texttt{over\_inference},
\texttt{stale\_memory\_use},
\texttt{privacy\_refusal},
\texttt{privacy\_overuse},
\texttt{attribution\_confusion},
\texttt{identity\_mis\_inference},
\texttt{incomplete\_memory\_use}, and
\texttt{false\_memory\_use}.

\section{Efficiency Analysis}
\label{app:efficiency}

We analyze the computational efficiency of ConMem on the 1,079 instances of the primary benchmark. Unless otherwise stated, the reported latency corresponds to the elapsed time of successful API requests. Failed attempts and retry backoff time are not included in the recorded latency. We report only the calls required by the primary ConMem setting and exclude additional calls performed solely for parameter-sensitivity analysis.

\paragraph{API Calls.}
The standard execution path requires 14 model calls per candidate memory: one call for future-probe generation, one for probe validation, six for generating the control and treatment responses to the three probes, three for blind response-quality evaluation, and three for memory-attribution auditing. A zero CMTE score introduces one additional tie-break call. In the primary evaluation, 199 of 1,079 cases (18.44\%) invoke this tie-break.

Probe validation can introduce additional calls when an initially generated probe set fails semantic validation. Across the full benchmark, ConMem performs 15,430 model-stage calls, corresponding to an average of 14.30 calls per case. Of these, 125 calls are additional probe-generation, validation, or repair calls beyond the standard execution path.

\paragraph{Latency.}
Table~\ref{tab:efficiency} reports latency statistics for each component. Response generation takes approximately 21 seconds per call on average for both the control and treatment branches. The attribution auditor is the most expensive repeatedly invoked component, with a mean latency of 26.85 seconds and a 95th-percentile latency of 85.76 seconds. Probe validation also exhibits a relatively long tail, with a 95th-percentile latency of 104.52 seconds.

\begin{table}[t]
\centering
\small
\setlength{\tabcolsep}{3.5pt}
\caption{Latency statistics of ConMem on the primary benchmark. All latency values are in seconds. ``Per-case'' denotes the sum of recorded API latencies required to evaluate one candidate memory.}
\label{tab:efficiency}
\begin{tabular}{@{}lrrrr@{}}
\toprule
Stage & Calls & Mean & Median & P95 \\
\midrule
Probe generation     & 1,129 & 26.76 & 25.19 & 41.63 \\
Probe audit          & 1,139 & 40.94 & 31.93 & 104.52 \\
Probe repair         & 15    & 17.00 & 15.96 & 23.49 \\
Control response     & 3,237 & 21.14 & 15.63 & 54.46 \\
Treatment response   & 3,237 & 21.72 & 16.88 & 53.95 \\
Quality judge        & 3,237 & 20.31 & 18.72 & 36.87 \\
Attribution auditor  & 3,237 & 26.85 & 15.94 & 85.76 \\
Tie-break judge      & 199   & 33.29 & 22.43 & 94.43 \\
\midrule
Per-case aggregate   & 1,079 & 347.64 & 325.44 & 577.46 \\
\bottomrule
\end{tabular}
\end{table}

The mean summed API latency is 347.64 seconds per case, with a median of 325.44 seconds and a 95th percentile of 577.46 seconds. These values represent the sum of component-level request latencies rather than the wall-clock duration of the complete experiment. In our implementation, up to 16 cases are processed concurrently; therefore, summed per-case latency should not be interpreted as total experimental runtime.

\paragraph{Probe-Generation Overhead.}
Probe generation is stable for most cases. A total of 1,044 cases (96.76\%) pass probe generation and semantic validation in the first round. Overall, 1,074 cases (99.54\%) are resolved through ordinary full-set generation or regeneration without entering targeted repair. Only four cases require targeted role repair, and one case requires fresh rescue synthesis.

\paragraph{Retry Behavior.}
Among the 15,430 model-stage calls, 15,334 (99.38\%) succeed on the first API attempt. Only 96 calls require at least one retry. The mean number of API attempts per model-stage call is 1.009, and the maximum observed number of attempts is five. Thus, transient API failures contribute little additional invocation overhead in the reported experiment.

These measurements reflect the API environment used in our experiments and should therefore be interpreted as implementation-level efficiency measurements rather than hardware-normalized inference benchmarks. Token usage was not logged during the original experiment and is consequently not reported.

\section{Backbone Robustness}
\label{app:backbone}

We further evaluate ConMem using DeepSeek-V4-Flash and compare it with GPT-5.6-sol on the same 1,079 candidate memories. Both backbones use three future-probe types, identical ConMem scoring coefficients, and the same score-first admission rule. Each backbone independently generates and evaluates its probes, allowing us to examine the end-to-end sensitivity to model selection.

\paragraph{Admission Performance.}
Table~\ref{tab:backbone} reports admission performance across the two backbones. GPT-5.6-sol achieves higher accuracy (77.85\% vs. 76.83\%), balanced accuracy (70.93\% vs. 67.76\%), and Macro-F1 (71.54\% vs. 68.89\%). The difference is mainly reflected in
Do-Not-Store recall (55.33\% vs. 47.33\%), whereas
DeepSeek-V4-Flash achieves slightly higher Store recall
(88.19\% vs. 86.52\%).

\begin{table}[!htbp]
\centering
\small
\setlength{\tabcolsep}{5pt}
\caption{ConMem admission performance across LLM backbones.
All values are percentages except counts.}
\label{tab:backbone}
\begin{tabular}{@{}lrr@{}}
\toprule
Metric & GPT-5.6 & DeepSeek-V4-Flash \\
\midrule
Accuracy          & 77.85 & 76.83 \\
Balanced Acc.     & 70.93 & 67.76 \\
Macro-F1          & 71.54 & 68.89 \\
\midrule
Store Precision   & 83.42 & 81.30 \\
Store Recall      & 86.52 & 88.19 \\
Store F1          & 84.94 & 84.61 \\
\midrule
DNS Precision     & 61.25 & 60.68 \\
DNS Recall        & 55.33 & 47.33 \\
DNS F1            & 58.14 & 53.18 \\
\midrule
Zero-score cases  & 458 & 193 \\
Zero-score rate   & 42.45 & 17.89 \\
\bottomrule
\end{tabular}
\end{table}

The frequency of zero-score cases also differs substantially:
42.45\% for GPT-5.6-sol versus 17.89\% for DeepSeek-V4-Flash.
Thus, the frequency of inconclusive counterfactual scores depends
on the backbone, although fewer zero-score cases do not necessarily
imply better admission performance.

\paragraph{Parameter Sensitivity.}
We examine 27 pre-specified combinations of three CMTE coefficients:
the irrelevant-probe mild-use penalty $\{0.25,0.5,0.75\}$,
the risk-probe safe-utility gain $\{0,0.25,0.5\}$,
and the risk-probe harm penalty $\{1.0,1.5,2.0\}$.
The primary setting remains fixed at $(0.5,0.5,1.5)$.

As shown in Table~\ref{tab:backbone_sensitivity},
GPT-5.6-sol achieves 77.85--79.43\% accuracy and
71.54--72.98\% Macro-F1 across these settings, compared with
76.65--77.94\% accuracy and 68.72--69.77\% Macro-F1 for
DeepSeek-V4-Flash. Both backbones exhibit relatively limited
variation within the tested coefficient ranges. These results
characterize parameter sensitivity rather than select an
alternative configuration.

\begin{table}[t]
\centering
\small
\setlength{\tabcolsep}{4.5pt}
\caption{Performance ranges across 27 CMTE scoring settings.
All values are percentages.}
\label{tab:backbone_sensitivity}
\begin{tabular}{@{}lrr@{}}
\toprule
Metric & GPT-5.6 & DeepSeek-V4-Flash \\
\midrule
Accuracy
 & 77.85--79.43 & 76.65--77.94 \\
Balanced Acc.
 & 70.93--71.92 & 67.63--68.33 \\
Macro-F1
 & 71.54--72.98 & 68.72--69.77 \\
Store Recall
 & 86.52--88.83 & 87.93--89.99 \\
DNS Recall
 & 54.67--55.67 & 46.67--47.33 \\
Zero-score Rate
 & 42.45--45.41 & 17.79--20.67 \\
\bottomrule
\end{tabular}
\end{table}

\end{document}